\documentclass[runningheads]{llncs}
\usepackage{graphicx}
\usepackage{amsmath}
\begin{document}
\title{HPointLoc: Point-based Indoor Place Recognition using Synthetic RGB-D Images\thanks{Supported in part of HPointLoc dataset preparation by a grant of Russian Science Foundation No. 20-71-10116 and in part of PNTR framework development by a grant for research centers in the field of artificial intelligence, provided by the Analytical Center for the Government of the Russian Federation in accordance with the subsidy agreement (agreement identifier 000000D730321P5Q0002) and the agreement with the Moscow Institute of Physics and Technology dated November 1, 2021 No. 70-2021-00138.}}
\titlerunning{HPointLoc}
%
\author{Dmitry Yudin\inst{1,4}\orcidID{0000-0002-1407-2633} \and
Yaroslav Solomentsev\inst{1,2}\orcidID{10000-0002-2664-7086} \and
Ruslan Musaev\inst{1}\orcidID{0000-0002-0679-7810}\and
Aleksei Staroverov\inst{1,4}\orcidID{0000-0002-4730-1543}\and
Aleksandr~I.~Panov\inst{1,3,4}\orcidID{0000-0002-9747-3837}}
\authorrunning{D. Yudin et al.}
%
\institute{Moscow Institute of Physics and Technology, Moscow Region, Dolgoprudny, 141700, Russia \and
LLC Integrant, Moscow, 127495, Russia
\and
Federal Research Center ``Computer Science and Control'' of the Russian Academy of Sciences, Moscow, 117312, Russia\\\and
AIRI (Artificial Intelligence Research Institute), Moscow, 105064, Russia\\
\email{yudin.da@mipt.ru, \{solomentsev.yak,musaev.rv\}@phystech.edu, alstar8@yandex.ru, panov.ai@mipt.ru}}
\maketitle              
\begin{abstract}

We present a novel dataset named as HPointLoc, specially designed for exploring capabilities of visual place recognition in indoor environment and loop detection in simultaneous localization and mapping. 
The loop detection sub-task is especially relevant when a robot with an on-board RGB-D camera can drive past the same place (``Point") at different angles.
The dataset is based on the popular Habitat simulator, in which it is possible to generate photorealistic indoor scenes using both own sensor data and open datasets, such as Matterport3D.
To study the main stages of solving the place recognition problem on the HPointLoc dataset, we proposed a new modular approach named as PNTR. It first performs an image retrieval with the Patch-NetVLAD method, then extracts keypoints and matches them using R2D2, LoFTR or SuperPoint with SuperGlue, and finally performs a camera pose optimization step with TEASER++. Such a solution to the place recognition problem has not been previously studied in existing publications.
The PNTR approach has shown the best quality metrics on the HPointLoc dataset and has a high potential for real use in localization systems for unmanned vehicles.
The proposed dataset and framework are publicly available: https://github.com/metra4ok/HPointLoc.

\keywords{Visual Place Recognition, Indoor Localization, Synthetic Image, RGB-D Image, Deep Learning, Dataset}
\end{abstract}
\section{Introduction}

Place recognition based on camera images is an important task for navigation of a robot or an unmanned vehicle \cite{staroverov2020real}.
This can significantly reduce the cost and simplify the determining of the agent spatial pose in a 3D scene.

Generally, visual localization has three key stages.
The first stage is retrieving for a given query image the most similar image from a previously known database images based on global embeddings \cite{netvlad,galvez2012bags,apgem}.
At the second stage, key points are extracted on the query image and on the retrieved image and matching them. Some modern methods combine these procedures into a single model \cite{loftr}.
Camera pose optimization is performed in the third step so that key points on the retrieved image coincide in location with the key points on the query image. Sometimes this is done in 2D \cite{g2o,ransac}, sometimes directly in 3D \cite{yang2020teaser,icp}. 

The modern trend is learning neural network-based methods that allow these stages to be performed. For them, it is important to have a diverse and large-scale dataset. It should contains, in addition to images, information about the exact camera position with 6 degrees of freedom (6DoF), and also data about distances corresponding to each pixel (depth map).

\begin{figure}[t]
      \centering
      \includegraphics[width=0.6\textwidth]{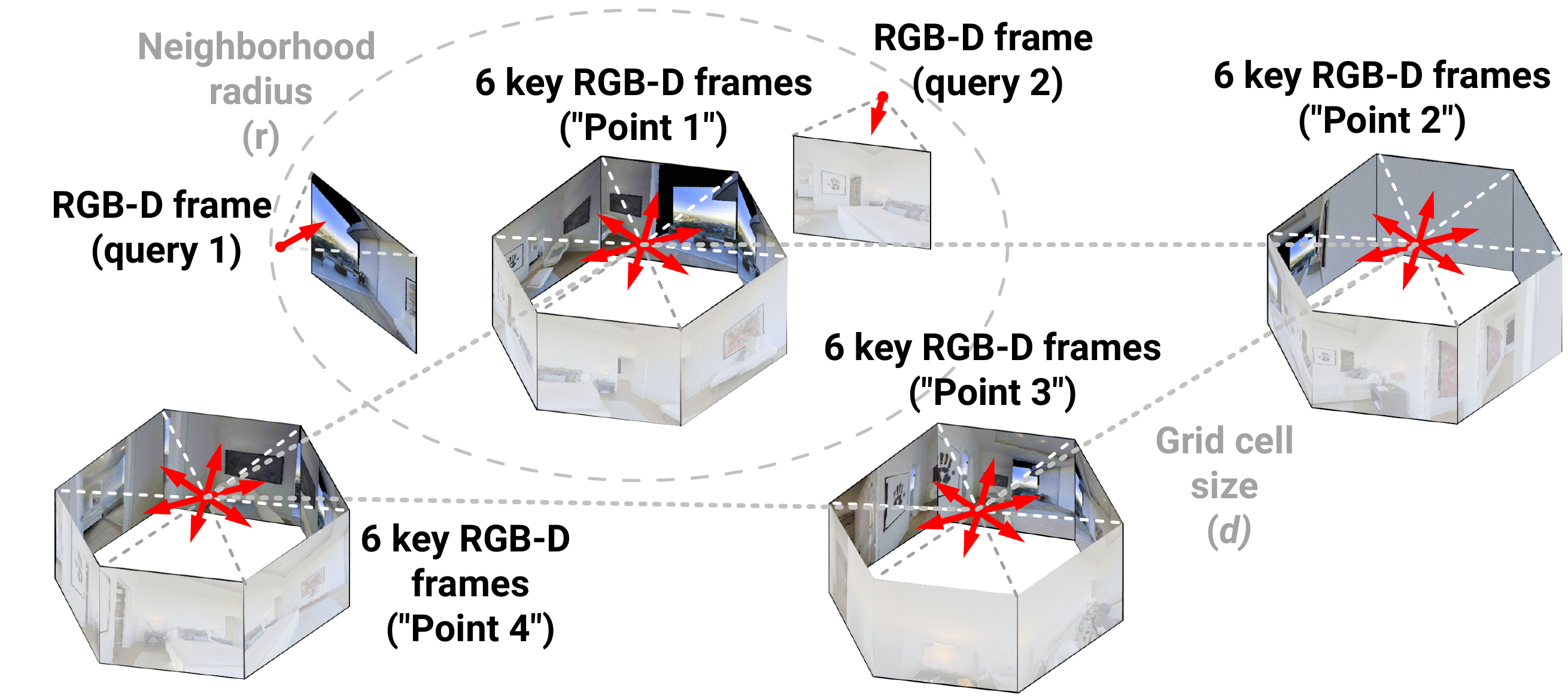}
      \caption{Illustration of the localization problem by RGB-D query image. It is necessary to determine the pose of the corresponding camera relative to the camera poses from known database, consisting of a regular grid of groups of 6 cameras (``Points")}
      \label{fig:task_statement}
   \end{figure}

This paper focuses on the development of the dataset for indoor localization that can later be used for robot's intelligent navigation in the photorealistic simulator Habitat \cite{habitat19iccv}. This will help to investigate quantitatively and qualitatively loop detection methods of a robot's movement when it enters the vicinity already visited place (``Points", see Fig.~\ref{fig:task_statement}).

Another contribution is the development of a new  modular approach named as PNTR.
It first performs an image retrieval with the Patch-NetVLAD method,
then extracts keypoints and matches them using R2D2, LoFTR or SuperPoint with SuperGlue, and finally performs a camera pose optimization
step with TEASER++. 

\section{Related Work}
\label{related}

\textbf{Image Retrieval.}
The search problem for the closest image in the database can be reformulated as a ranking problem. The solution requires finding informative and compact local and global descriptors of the images. 

The common ``classical" approaches obtain global features (embeddings) by aggregating the local descriptors using the bag of words (BoW) scheme (DBoW2, DBoW3, FBoW) or vectors of locally aggregated descriptors (VLAD). 

In the last few years, new approaches based on deep neural networks have been released: NetVLAD\cite{netvlad}, distilled model HF-Net\cite{hierarchical_localization}, Ap-Gem\cite{apgem} approach with differentiable rank loss function, 
graph-based approach GraphVLAD\cite{zhang2021lifted}. They had surpassed the classical ones by feature learning for the specific problems.
The similar image candidates can be also re-ranked by analyzing statistics of geometrically correct matchings of local descriptors for image patches as in Patch-NetVLAD\cite{hausler2021patchnetvlad}. 

To improve the matching of embeddings, some approaches utilize semantic information. They demonstrate good performance on popular benchmarks \cite{neubert2021vector,peng2021semantic,xue2022efficient}.
However, in spite of notable achievements of these neural networks, there still is a problem with the extraction of invariant semantic descriptors of images which have only low-quality semantic/instance segmentation. Such segmentation is often inherent in real-time neural networks that generate a lot of noise in the resulting masks. 

\textbf{Local Feature Extraction and Matching.}
In the next stage of visual localization, we should find (detect and describe) common local features (keypoints) on pairs of images. The difficulties of this problem include getting informative and compact local descriptors and providing geometrical verification of detected keypoints. 

The classical local features extraction approaches utilize scale-invariant feature transform based on local gradients \cite{sift}, BoW \cite{galvez2012bags}, etc. These methods have achieved remarkable performance, but analogously to image retrieval methods, modern neural network approaches have outperformed them by finding more robust local features. The recent approaches demonstrate good results due to appropriate loss functions, neural network architectures (CNN, Transformers), and training schemes (Siamese networks). For example, the advantage of end-to-end training allows us to simultaneously train keypoint detectors and descriptors: SuperPoint \cite{superpoint}, R2D2 \cite{r2d2}, D2-Net \cite{dusmanu2019d2}, etc. 

The goal of the feature matching is to find the same keypoint descriptors in both query and retrieved images. Metric learning is the most common training scheme for this purpose. Models using attention mechanisms (LoFTR \cite{loftr}) and graphs (SuperGlue \cite{superglue}) provide highly accurate keypoint matching. In the inference mode, models can use the k-nearest neighbors algorithm for fast and robust matching (Faiss\cite{JDH17}). 

\textbf{Camera Pose Optimization}
is the final stage of the visual localization. The main problem is to find such transformation (SE3 rotation matrix and translation vector), which minimizes the pairwise distance between point clouds. The most popular methods for this task are the classical optimization approaches such as easy PnP + Ransac\cite{ransac}, graph-based g2o\cite{g2o}, 3D point cloud-based ICP\cite{icp}, reliable to outliers TEASER++\cite{yang2020teaser}, neural network-based methods PoseNet\cite{kendall2015posenet}, DCP \cite{wang2019deep}, etc.

\textbf{Complex Localization Approaches.}
One of the most popular three-stage approach using neural networks is Hierarchical-Localization \cite{hierarchical_localization}. HF-Net\cite{hierarchical_localization} is an image retrieval method in this pipeline, SuperPoint\cite{superpoint} is a keypoint extracting method, SuperGlue\cite{superglue} is used for keypoint matching and PnP\cite{pnp} with RANSAC\cite{ransac} are for pose optimization.
Another state-of-the-art method on long-term visual localization benchmark\cite{cvpr} is Kapture framework using Ap-GeM \cite{apgem} and R2D2 \cite{r2d2}. 

One of the most popular classic non-neural network visual localization method is the combination of ORB\cite{orb} and DBoW2\cite{galvez2012bags}, which is widely used in popular SLAM methods, in particular, ORB-SLAM2, OpenVSLAM, etc. 

\begin{table*}[t]
\centering
\tiny
 \caption{
 Open datasets for visual place recognition and localization
 }
    \label{tab:datasets}
    \begin{tabular}{|l||l|l|l|p{6em}|l|p{8em}|p{2em}|p{2em}|p{8em}|}
    \hline
    Dataset          & Year & Scene & Synth. & Pose info. & RGB  & Depth info. & Sem. segm. & Inst. segm. & Frames                  \\
    \hline
    Pittsburgh 
    \cite{masone2021survey} & 2015 & outdoor    &           & GPS              & $\times$         &                               &                       &                       & 278k                              \\\hline
    Landmarks 
    \cite{masone2021survey} & 2016 & outdoor    &           & Label            & $\times$        &                               &                       &                       & 10k                               \\\hline
    Google-Landmarks 
    \cite{masone2021survey} & 2017 & outdoor    &           & GPS              & $\times$         &                               &                       &                       & 1.2M                              \\\hline
    Nordland 
    \cite{masone2021survey} & 2018 & outdoor    &           & GPS              & $\times$         &                               &                       &                       & 143k                              \\\hline
    CMU-Seasons 
    \cite{cvpr} & 2018 & outdoor    &           & 6DoF Pose        & $\times$         & laser scan.                 &                       &                       & 82.5k                               \\\hline
    Aachen Day-Night 
    \cite{cvpr} & 2018 & outdoor    &           & 6DoF Pose        & $\times$         & stereo                        &                       &                       & 7.5k                               \\\hline 
    RobotCar Seasons
    \cite{cvpr}  & 2018 & outdoor    &           & 6DoF Pose        & $\times$         & laser scan.                 &                       &                       & 38k                               \\\hline
    Tokyo 24/7  
    \cite{masone2021survey} & 2018 & outdoor    &           & GPS              & $\times$         & stereo                        &                       &                       & 2.8M                              \\\hline
    Argoverse Stereo  \cite{chang2019argoverse}        & 2019 & outdoor    &           & 6DoF Pose        & $\times$         & stereo/laser scan.        &                       &                       & 6,6k                              \\\hline
    NuScenes-lidarseg  \cite{nuscenes2019}       & 2020 & outdoor    &           & 6DoF Pose        & $\times$         & laser scan.                 & $\times$                     &                       & 40k point clouds \\\hline
    Waymo Perception \cite{sun2020scalability}  & 2020 & outdoor    &           & 6DoF Pose        & $\times$         & stereo/laser scan.        &                       &                       & 390k                              \\\hline
    KITTI360 \cite{Xie2016CVPR}              & 2020 & outdoor    &           & 6DoF Pose        & $\times$         & laser scan.                 & $\times$                     & $\times$                     & 4$\times$83k                             \\\hline
    Mapillary SLS 
    \cite{masone2021survey} & 2020 & outdoor    &           & 6DoF Pose        & $\times$         &                               &                       &                       & 1.68M                             \\\hline
    TUM Indoor 
    \cite{lee2021large} & 2012 & indoor     &           & 6DoF Pose        & $\times$         & laser scan.                 &                       &                       & 7k                                \\\hline
    7-scenes 
    \cite{masone2021survey} & 2013 & indoor     &           & 6DoF Pose        & $\times$         & RGB-D cam.                 &                       &                       & 17k                               \\\hline
    Baidu 
    \cite{yu2017baidu} & 2017 & indoor     &           & 6DoF Pose        & $\times$         & laser scan.                 & $\times$                     &                       & 2k                                \\\hline
    
    Matterport3D \cite{chang2017matterport3d}             & 2017 & indoor     & $\times$         & Need to gen. & $\times$         & 3D Models & $\times$                    & $\times$                     & 194k                              \\\hline
    TUM-LSI  
    \cite{lee2021large} & 2017 & indoor     &           & 6DoF Pose        & $\times$         & laser scan.                 &                       &                       & 220                               \\\hline
    ScanNet  
    \cite{lee2021large} & 2017 & indoor     &           & 6DoF Pose        & $\times$         & RGB-D cam.                  &    $\times$                    & $\times$                    & 2.4M                              \\\hline
    2D-3D-Semantics \cite{2017arXiv170201105A}          & 2017 & indoor     &           & 6DoF Pose        & $\times$        & RGB-D cam.     & $\times$                     &                       & 70k                               \\\hline
    Gibson \cite{xiazamirhe2018gibsonenv}                   & 2017 & indoor     & $\times$         & Need to gen. & $\times$         & 3D Models &                       &                       & 572 3D Scenes                     \\\hline
    Inloc  
    \cite{cvpr} & 2018 & indoor     &           & 6DoF Pose        & $\times$         & 3D point clouds&                       &                       & 10k                               \\\hline
    Replica   \cite{replica19arxiv}                & 2019 & indoor     & $\times$         & Need to gen. & $\times$         & 3D Models & $\times$                     & $\times$                     & 18 3D Scenes                      \\\hline
    ROI10   \cite{wald2020beyond}                    & 2020 & indoor     &          & 6DoF Pose & $\times$         & RGB-D cam. &                       &                       & 250k                    \\\hline
    HM3D  \cite{habitat_mp3d1}                    & 2021 & indoor     & $\times$         & Need to gen. & $\times$        & 3D Models &                       &                       & 1000 3D scenes                    \\\hline
    Naver Labs 
    \cite{lee2021large} & 2021 & indoor     &           & 6DoF Pose        & $\times$        & laser scan.                 & $\times$                     &                       & 100k                                \\\hline
    HPointloc (our)          & 2022 & indoor     & $\times$        & 6DoF Pose        & $\times$         & RGB-D cam.                  & $\times$                     & $\times$                     & 76k   \\  \hline                      
    \end{tabular}
\end{table*}

\textbf{Datasets.}
There are many known datasets used to solve the visual localization problem (see Table \ref{tab:datasets}). The most popular of them are presented in the long-term visual localization challenge \cite{cvpr}, as well as in the survey publications \cite{lee2021large,masone2021survey}. 
One of the drawbacks of most of them is the lack of data on instance or semantic segmentation, or depth maps, while this information is very important for improving existing localization approaches.

If we take the InLoc dataset \cite{taira2018inloc} as an example, then it includes not dense depth maps, but sparse point clouds. Using them we can obtain an estimate of the depths for the found keypoints, but in our formulation of the problem, we do not consider such a case.

Of particular note are datasets that contain 3D models of a scene, rather than separate frames (Matterport3D\cite{chang2017matterport3d}, Gibson\cite{xiazamirhe2018gibsonenv}, Replica\cite{replica19arxiv}, HM3D\cite{habitat_mp3d1}). They are usually integrated into simulators of the movement of intelligent agents, for example, into the popular and computationally efficient Habitat\cite{habitat19iccv}. At the same time, to obtain a reproducible and useful result based on them, it is necessary to have sufficiently large and diverse samples from these datasets containing frames and the corresponding 6DoF camera poses.

In our work, we propose filling the gap among such datasets, specialized in the study of a specific problem — localization of an intelligent agent in the vicinity of some key point belonging to a regular grid in a 3D indoor environment. 

\begin{figure*}[t]
\centering
\includegraphics[width=0.9\textwidth]{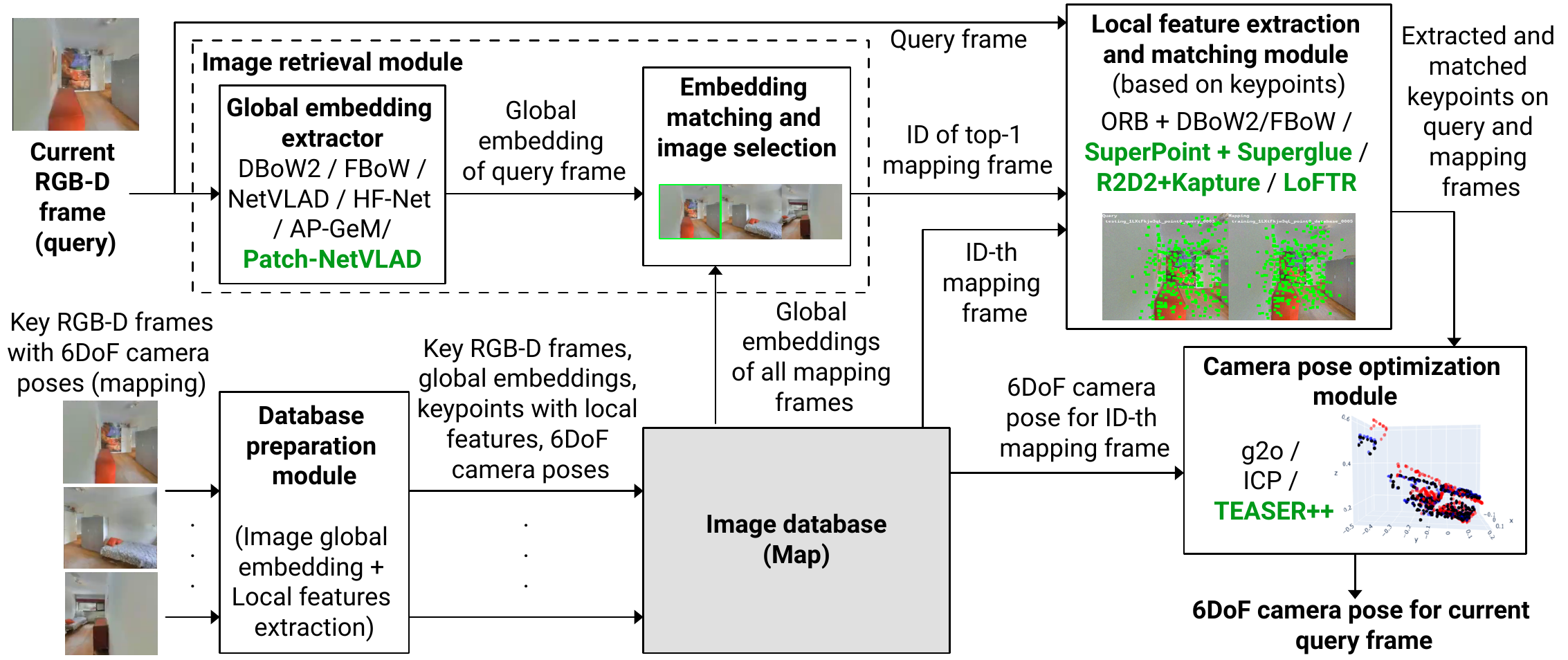}
\caption{
Scheme of the proposed framework for indoor localization using RGB-D images. 
Green text refers to the methods used in the PNTR approach.
}
\label{fig:pipeline}
\end{figure*}

\section{The Point-based Approach to Indoor Place Recognition}
\label{problem}
In this paper, we solve the problem of developing an approach to estimating the 6DoF pose $P_{q}$ of an intelligent agent from the image $I_{q}$ of its RGB-D camera (query) in the vicinity of the poses of the cameras $P^{i}_{db}$ from the database (see Figure \ref{fig:task_statement}), forming a regular grid of groups of 6 cameras (named as ``Points"), the images $I^{i}_{db}$ of which cover 360$^{\circ}$ environment.
This formulation required the development of a new special dataset, named as HPointLoc. The dataset was generated automatically based on the Habitat simulator\cite{habitat19iccv}. This will provide the perspective to use the results for navigation and learning the behavior of an intelligent agent (robot) in real time in this photorealistic environment.
On such a ``Point"-based dataset, it is advisable to evaluate the quality of the state-of-the-art methods of visual place recognition. For their study, we proposed the modular PNTR approach, which is based on trainable methods. Its modules are described in more detail later in the paper. 

\textbf{Image Retrieval Module.}
When developing the approach, various classical image retrieval models based on a bag of words and neural network approaches are investigated. They form global feature vectors (embeddings) $E_{q}$ and $E^{i}_{db}$ for the query image $I_{q}$ and images from the database $I^{i}_{db}$. Based on embeddings, the most similar image in the sense of the similarity metric $S$ is determined:

\begin{equation}
top1 = \underset{i}{\operatorname{argmin}}\left(S\left( E_{q}, E^{i}_{db}\right)\right).
\label{eq:retrieval}
\end{equation}

We consider the classic approaches DBoW2 \cite{galvez2012bags} and FBoW \cite{zhao2017fuzzy} based on the formation of global descriptors by image keypoints extracted with ORB method. The main attention is paid to neural network-based approaches of image retrieval: NetVLAD\cite{netvlad}, AP-GeM\cite{apgem}, HF-Net\cite{hierarchical_localization} and the Patch-NetVLAD\cite{hausler2021patchnetvlad} (its speed up configuration (s) is included in our PNTR approach). 

\textbf{Local Feature Extraction and Matching Module.}
At the next stage of extraction and matching of local features, $k$ matches of keypoints on the query image $F^{j}_{q}$ and the most similar from the base $F^{j}_{top1}$ are found, $j=1..k$. For each of these points, the distance to the camera in meters (depth) is known: $D^{j}_{q}$ and $D^{j}_{top1}$, $j=1..k$ respectively.

We examined the extraction and matching of image keypoints based on the classic popular approaches ORB with DBoW2 \cite{galvez2012bags} or FBoW \cite{zhao2017fuzzy} and the popular neural networks (used in our PNTR approach) SuperPoint \cite{superpoint} with SuperGlue \cite{superglue}, R2D2 \cite{r2d2} with feature matching based on the Kapture toolbox, as well as monolithic LoFTR \cite{loftr}.

\textbf{Database Preparation Module.}
The database used to solve the localization problem is an array of $i\in[1, N]$ camera poses $P^{i}_{db}$, the corresponding images $I^{i}_{db}$, depth maps $D^{i}_{db}$, and segmentation masks $M^{i}_{db}$ (the latter may not be used in the approach). In addition, extracted and pre-calculated global image embeddings $E^{i}_{db}$ and information about local features $F^{j}_{db}$ on them are additionally entered into the database.

\textbf{Camera Pose Optimization Module.}
Further, the final estimation of the 4$\times$4 matrix of the camera pose $P_{q}$ is carried out.
This pose corresponds to the query image $I_{q}$. In our case, the 4$\times$4 matrix of the camera pose $P^{top1}_{db}$ is known for retrieved image $I^{top1}_{db}$ from database. 
An optimization problem is solved for calculating the relative pose of the camera $P^{q}_{top1}$ (also a 4$\times$4 matrix) based on minimizing some functional $L$ that takes into account the 2D coordinates $F$ and the depth $D$ of matched keypoints in two images:
\begin{equation}
P^{q}_{top1} = \underset{P^{q}_{top1},j\in[1,k]}{\operatorname{argmin}}\left(L\left( F^{j}_{q},D^{j}_{q},F^{j}_{top1},D^{j}_{top1}\right)\right).
\label{eq:pose}
\end{equation}
The final pose is defined as $P_{q} = P^{top1}_{db} P^{q}_{top1}.$

The widely used classical approaches g2o\cite{g2o}, ICP\cite{icp} and the newer and highly accurate Teaser++\cite{yang2020teaser} (used in the proposed PNTR approach) are investigated as methods for optimizing pose in our framework.

To quantitatively evaluate the quality of the overall framework, we use Recall metric with different thresholds. It is calculated as the fraction of query images which localization errors do not exceed the specified threshold, respectively for distance (translation) $\epsilon_{t}\in\{0.25m, 0.5m, 1m, 5m\}$ and rotation $\epsilon_{r}\in\{2^{\circ}, 5^{\circ}, 10^{\circ}, 20^{\circ}\}$. Such thresholds were chosen to assess the prospects for solving the problem of loop detection and place recognition in selected indoor environment using the developed approach.

\section{HPointLoc Dataset}
\label{dataset}

The dataset consists of 49 scenes from the Matterport3D dataset and is intended for training computer vision algorithms or testing them.
It was formed according to the following algorithm. 

\begin{figure*}[t]
    \centering
    \includegraphics[width=1.0\textwidth]{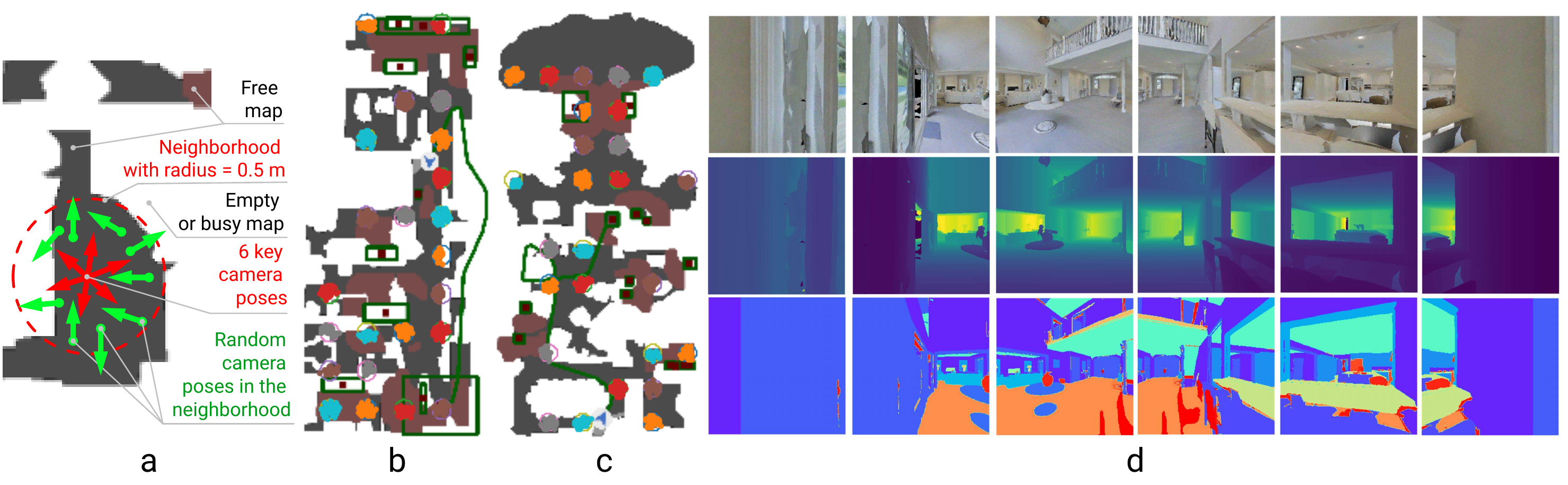}
    \caption{Explanations to the HPointLoc dataset content: a - the generation of query and database camera poses based on maps from the Matterport3D dataset in the Habitat environment; b - scene map in the HPointLoc-Val dataset; c - another scene map in the HPointLoc-All dataset, the positions of the camera centers are shown by colored dots in the circles corresponding to the ``Points"; d - examples of 6 RGB-images (upper row), depth map (middle row) and map of object instance segmentation (lower row), corresponding to one ``point" of scene map of the HPointLoc dataset. Six RGB-images from the ``point" cover overlapping areas with a 360-degree view}
    \label{fig:map_point}
\end{figure*}

For each scene, a regular grid with $N$ key poses was generated $({x}_{id},{y}_{id},{z}_{id}$, ${yaw}_{id}, {pitch}_{id},{roll}_{id})$ with a distance of 2 meters in $x$ and $y$ axes for one scene.
A key pose has a unique ${id}$ and field ${status=1}$. An example of a key pose is marked in red in Fig. \ref{fig:map_point}a. For each key pose, the two steps were taken:

1. Generation of five more key pose orientations with 60 degrees step, simulating a 360 degree key pose camera. The view angle of each frame is 90 degrees. An example of six key images is shown in Fig.~\ref{fig:map_point}d.

2. Generation of $M=50$ random poses in a radius $r=0.5m$ relative to the key pose $({x}_{id},{y}_{id},{z}_{id})$ with a random orientation (shown in green in Fig.~\ref{fig:map_point}a). If the generated pose does not fall into the free area, then it is discarded; otherwise we add a new record to the dataset.

\begin{table}[t]
  \centering
  \scriptsize
  \caption{Summary for HPointLoc Dataset.
  }
    \begin{tabular}{|l||l|l|l|l|l|}
    \hline
    & Points & Poses & Categories & Instances & Maps \\
    \hline
    HPointLoc-Val & 23    & 1088  & 33    & 3266  & 1 \\
    HPointLoc-All & 1757  & 86678 & 41    & 488717 & 49 \\
    \hline
    \end{tabular}%
  \label{tab:dataset_short_statistics}%
\end{table}%

\begin{table}[t]
    \caption{Statistics for semantic instances in the proposed HPointLoc-Val and HPointLoc-All datasets}
    \centering
    \tiny
    \begin{tabular}{|l|l|l||l|l|l||l|l|l|}
    \hline
        &\multicolumn{2}{|l||}{Instances per categ.}&   & \multicolumn{2}{|l||}{Instances per categ.} &  & \multicolumn{2}{|l|}{Instances per categ.} \\ 
        Category & Val & All & Category & Val & All & Category & Val & All \\ \hline \hline
        appliances & 0 & 1077 & cushion & 15 & 5600 & shelving & 18 & 4174 \\ \hline
        bathtub & 11 & 621 & door & 1174 & 44500 & shower & 4 & 2190 \\ \hline
        beam & 0 & 981 & fireplace & 0 & 1022 & sink & 6 & 1924 \\ \hline
        bed & 28 & 5025 & floor & 0 & 52274 & sofa & 6 & 3871 \\ \hline
        blinds & 0 & 162 & furniture & 8 & 376 & stairs & 15 & 3557 \\ \hline
        board\_panel & 3 & 145 & gym equipment & 10 & 105 & stool & 0 & 7660 \\ \hline
        cabinet & 26 & 5383 & lighting & 184 & 8797 & table & 15 & 11886 \\ \hline
        ceiling & 249 & 28143 & mirror & 16 & 1752 & toilet & 3 & 271 \\ \hline
        chair & 48 & 27946 & misc & 180 & 47346 & towel & 17 & 482 \\ \hline
        chest\_of\_drawers & 8 & 2123 & objects & 67 & 28902 & tv\_monitor & 10 & 1190 \\ \hline
        clothes & 0 & 191 & picture & 115 & 18945 & void & 10 & 14445 \\ \hline
        column & 0 & 4255 & plant & 2 & 4330 & wall & 601 & 107291 \\ \hline
        counter & 0 & 1521 & railing & 0 & 8157 & window & 110 & 19041 \\ \hline
        curtain & 11 & 8129 & seating & 6 & 2927 & & & \\ \hline
    \end{tabular}
    \label{tab:dataset_instance_statistics}
\end{table}

Each item in the dataset contains the following data: RGB and depth images, instance segmentation of the frame with 41 classes (see Table~\ref{tab:dataset_instance_statistics}) and GPS with Compass data (orientation of the camera is described with a quaternion).

RGB images have 256$\times$256 size and camera viewing angle 90$^{\circ}$. Gaussian noise model was also added with factor intensity of 0.02, a mean value of 0 and a sigma parameter of 1. 

Depth images  also have 256$\times$256 size. The depth values lie in the range from 0 to 1, where 0 is a minimum possible depth (0 m) and 1 — maximum possible (10 m). Such maximum depth is chosen because most of the commercial RGB-D cameras, such as Intel RealSense, and ZED, have similar restrictions. The RGB-D camera stands at 1.25 meters high above the floor.

Examples of top-view scene maps with marked frame poses are shown in Fig. \ref{fig:map_point}b and \ref{fig:map_point}c.
Thus, in each spherical ``Point" with a radius of 0.5 meters, there are 6 key images with the prior known data for localization methods (images in the database) and several dozen images (query images), which need to be localized. It is assumed in the experiments that the pose of query images by which it is necessary to localize is not known to the localization method, but all other data, including the depth map, are known. A summary of the dataset contents is shown in Table \ref{tab:dataset_short_statistics}. The dataset is split into two parts: the validation HPointLoc-Val, which contains only one scene, and the complete HPointLoc-All dataset, containing all 49 scenes, including HPointLoc-Val. We use these datasets only to validate the proposed approaches, not for training. 

\section{Experimental Results}
\label{experimental_results}

\begin{table*}[t]
  \centering
  \tiny
  \caption{Localization quality based on the pose of top-1 retrieved image (without pose optimization) on the HPointLoc-Val dataset}
    \begin{tabular}{|p{11em}||p{8em}|llll|llll|}
    \hline
    Image retrieval method & Embedding size & (5m,20°) & (1m,10°) & (0.5m,5°) & (0.25m,2°) & (5m)    & (1m)    & (0.5m)  & (0.25m) \\
    \hline
    FBoW+ORB & $\sim$1000000 & 0.527   & 0.269   & 0.133   & 0.029   & 0.911   & 0.864   & 0.864   & 0.476 \\
    
    DBoW2+ORB & $\sim$1000000 & 0.542   & 0.269   & 0.133   & 0.028   & 0.916   & 0.877   & 0.877   & 0.477 \\
    
    AP-GeM & 2048 & 0.559   & 0.282   & 0.138   & \textbf{0.032}   & 0.964   & 0.893   & 0.893   & 0.493 \\
    
    NetVLAD & 32768 & 0.584   & 0.291   & \textbf{0.141}   & \textbf{0.032}   & 0.971   & 0.919   & 0.919   & 0.502 \\
    
    HF-Net & 4096 & \textbf{0.600}     & 0.291   & 0.138   & \textbf{0.032}   & 0.975   & 0.934   & 0.934   & 0.505 \\
    
    Patch-NetVLAD(s) & 512$\times$100 patches &
	0.590 & \textbf{0.296} &	0.140 & 0.030 &	 \textbf{0.983} &	 \textbf{0.963} &   \textbf{0.963} & 0.515\\
    
    Patch-NetVLAD(p) & 4096$\times$300 patches & 0.576   & 0.292   & 0.138   & 0.028   & 0.979   & 0.961   & 0.961   &   \textbf{0.517} \\
    
    \hline
    \end{tabular}%
  \label{tab:image_retrieval_methods_loc}%
\end{table*}%

\begin{table*}[t]
  \centering
  \tiny
  \caption{Quality of various visual localization methods on the HPointLoc-Val dataset}
    \begin{tabular}{|p{20em}||llll|llll|}
    \hline 
     Approach & (5m,20°) & (1m,10°) & (0.5m,5°) & (0.25m,2°) & (5m)    & (1m)    & (0.5m)  & (0.25m) \\
    \hline
    DBoW2+ORB+g2o &  \textbf{0.844}  &  \textbf{0.812}   &  \textbf{0.78}    &  \textbf{0.726}   &  \textbf{0.905 }  &  \textbf{0.857}   & \textbf{0.841}   &  \textbf{0.796} \\
    FBoW+ORB+g2o & 0.801   & 0.766   & 0.719   & 0.64    & 0.903   & 0.826   & 0.78    & 0.719 \\
    \hline
    
    AP-GeM+R2D2(K)+ICP &   \textbf{0.961}   & 0.862   & 0.805   & 0.733   &   \textbf{0.965}   & 0.871   & 0.828   & 0.777 \\
    AP-GeM+R2D2(K)+ TEASER++ & 0.939   &   \textbf{0.904}   &   \textbf{0.881}   &   \textbf{0.877}   & 0.958   &   \textbf{0.906}   & 0.882   &   \textbf{0.880} \\
    Ap-GeM+LofTR+g2o & 0.907   & 0.868   & 0.837   & 0.746   & 0.963   & 0.902   &   \textbf{0.891}   & 0.861 \\
    Ap-Gem+LofTR+ICP & 0.903   & 0.846   & 0.825   & 0.762   & 0.961   & 0.877   & 0.862   & 0.807 \\
    AP-GeM+LoFTR+ TEASER++ & 0.892   & 0.854   & 0.831   & 0.766   & 0.964   & 0.899   & 0.884   & 0.823 \\
    \hline
    
    NetVLAD+SP+SuperGlue+g2o & 0.917   & 0.891   & 0.874   & 0.845   & 0.967   & 0.919   & 0.906   & 0.887 \\
    NetVLAD+SP+SuperGlue+ICP  & 0.944   & 0.893   & 0.868   & 0.794   & 0.969   & 0.909   & 0.893   & 0.842 \\
    NetVLAD+SP+SuperGlue+TEASER++ & 0.924   & 0.892   & 0.872   & 0.854   & 0.966   & 0.918   & 0.906   & 0.888 \\
    NetVLAD+R2D2(K)+ICP &   \textbf{0.968}   & 0.901   & 0.855   & 0.788   & 0.969   & 0.907   & 0.877   & 0.827 \\
    NetVlad+R2D2+TEASER++ & 0.941 &   \textbf{0.916}   &   \textbf{0.907}   &   \textbf{0.905}   & 0.967   & 0.918   & 0.911   &   \textbf{0.908} \\
    NetVLAD+LoFTR+g2o & 0.921   & 0.891   & 0.876   & 0.78    & 0.968   &   \textbf{0.92}    &   \textbf{0.916}   & 0.892 \\
    NetVLAD+LoFTR+ICP  & 0.917   & 0.875   & 0.858   & 0.799   &   \textbf{0.971}   & 0.9     & 0.892   & 0.843 \\
    NetVLAD+LoFTR+ TEASER++ & 0.908   & 0.881   & 0.865   & 0.806   &   \textbf{0.971}   & 0.919   & 0.914   & 0.857 \\
    
    \hline
    HF-Net+SP+SuperGlue+g2o & 0.925   & 0.903   & 0.881   & 0.852   & 0.972   & 0.936   & 0.92    & 0.896 \\
    HF-Net+R2D2(K)+ICP &   \textbf{0.974}   & 0.917   & 0.869   & 0.791   &   \textbf{0.975}   & 0.925   & 0.894   & 0.834 \\
    HF-Net+R2D2(K)+ TEASER++ & 0.953   &   \textbf{0.936}   &   \textbf{0.923}   &   \textbf{0.92}    & 0.973   &   \textbf{0.938}   &   \textbf{0.926}   &   \textbf{0.923} \\
    
    \hline
    Patch-NetVLAD(s)+SP+SuperGlue+ g2o & 0.946   & 0.924   & 0.908   & 0.883   &   \textbf{0.982}   &   \textbf{0.96}    & 0.947   & 0.924 \\
    Patch-NetVLAD(s)+ R2D2(K)+ICP &   \textbf{0.98}    & 0.94    & 0.896   & 0.808   & 0.984   & 0.952   & 0.925   & 0.852 \\

    PNTR with SP+SuperGlue &0.947	&0.925	&0.899	&0.882	&0.975	&0.945	&0.934	&0.914\\
    PNTR with R2D2(K) & 0.964   &   \textbf{0.957}   &   \textbf{0.952}   & \textbf{0.945}   & 0.98    & 0.958   &   \textbf{0.953}   &   \textbf{0.951} \\
    PNTR with LoFTR &0.942	&0.919	&0.900	&0.834	&0.977	&0.953	&0.948	&0.892\\
    \hline
    \end{tabular}%
  \label{tab:pipeline_metrics_val}%
\end{table*}%

\begin{table*}[t]
  \centering
  \tiny
  \caption{Quality of various visual localization methods on the HPointLoc-All dataset}
    \begin{tabular}{|p{19em}||llll|llll|}
    \hline  
     Approach & (5m,20°) & (1m,10°) & (0.5m,5°) & (0.25m,2°) & (5m)    & (1m)    & (0.5m)  & (0.25m) \\
    \hline \hline
    
    DBoW2+ORB (top-1 db) & 0.592   & 0.303   & 0.150    & 0.033   & 0.903   & 0.882   & 0.881   & 0.498 \\
    FBoW+ORB2 (top-1 db) &	0.547 &	0.285 &	0.142 &	0.032 &	0.825 &	0.799 &	0.797 &	0.464 \\
    
    NetVLAD (top-1 db) & 0.643   & 0.317   & 0.158   &  \textbf{0.034}   & 0.957   & 0.877   & 0.876   & 0.487 \\
    
    AP-GeM (top-1 db) & 0.635   & 0.311   & 0.156   &  \textbf{0.034}   & 0.944   & 0.813   & 0.812   & 0.452 \\
    
    HF-Net (top-1 db) &  \textbf{0.646}   & 0.318   & 0.158   &  \textbf{0.034}   & 0.955   & 0.879   & 0.878   & 0.487 \\
    Patch-NetVLAD(s) (top-1 db) & 0.644 &  \textbf{0.320} &	 \textbf{0.159} &	 \textbf{0.034} &	 \textbf{0.969} &	 \textbf{0.944} &	 \textbf{0.942} &	 \textbf{0.516} \\
    
    \hline
    
    DBoW2+ORB2+g2o &  \textbf{0.876}   &  \textbf{0.857}   &  \textbf{0.648}   &  \textbf{0.404}   &  \textbf{0.901}   &  \textbf{0.870}    &  \textbf{0.661}   &  \textbf{0.419} \\
    FBoW+ORB2+g2o &	0.776 &	0.748 &	0.571 &	0.357 &	0.820 &	0.769 &	0.590 &	0.377 \\
    
    \hline
    NetVLAD+SP+SuperGlue+ g2o & 0.925   &  \textbf{0.867}   & 0.647   & 0.406   & 0.950    &  \textbf{0.877}   & 0.656   & 0.414 \\
    NetVLAD+SP+SuperGlue+ICP & 0.935   & 0.853   & 0.803   & 0.426   &  \textbf{0.956}   & 0.876   & 0.844   & 0.476 \\
    NetVLAD+R2D2(K)+ICP &  \textbf{0.944}   & 0.842   & 0.748   & 0.341   &  \textbf{0.956}   &  \textbf{0.877}   & 0.846   & 0.474 \\
    NetVLAD+LoFTR+ICP & 0.934   & 0.863   &  \textbf{0.829}   &  \textbf{0.458}   &  \textbf{0.956}   &  \textbf{0.877}   &  \textbf{0.849}   &  \textbf{0.480} \\
    NetVLAD+LoFTR+TEASER++ & 0.933   & 0.865   & 0.829   & 0.445   &  \textbf{0.956}   &  \textbf{0.877}   &  \textbf{0.849}   & 0.479 \\
    
    \hline
    Ap-Gem+LofTR+ICP &  \textbf{0.907}   & 0.797   &  \textbf{0.765} &  \textbf{0.423}   &  \textbf{0.943} &  \textbf{0.813}   &  \textbf{0.787}   &  \textbf{0.445} \\
    AP-GeM+LoFTR+ TEASER++  &  \textbf{0.907}   &  \textbf{0.799} &  \textbf{0.765}   & 0.412   &  \textbf{0.943} &  \textbf{0.813}   &  \textbf{0.787}   &  \textbf{0.445} \\
    
    \hline
    HF-Net+SP+SuperGlue+g2o & 0.924   &  \textbf{0.868}   & 0.647   & 0.405   & 0.949   & 0.878   & 0.656   & 0.413 \\
    HF-Net+SP+SuperGlue+ICP & 0.94    & 0.861   & 0.817   & 0.439   &  \textbf{0.955}   &  \textbf{0.879}   & 0.85    & 0.479 \\
    HF-Net+SP+SuperGlue+TEASER++ & 0.935   & 0.854   & 0.803   & 0.425   &  \textbf{0.955}   & 0.877   & 0.845   & 0.475 \\
    HF-Net+R2D2(K)+ICP &  \textbf{0.943}   & 0.845   & 0.752   & 0.342   &  \textbf{0.955}   &  \textbf{0.879}   & 0.848   & 0.475 \\
    HF-Net+LoFTR+TEASER++  & 0.933   & 0.867   &  \textbf{0.831}   &  \textbf{0.445}    &  \textbf{0.955}   &  \textbf{0.879}   &  \textbf{0.851}   &  \textbf{0.480} \\
    \hline
    Patch-NetVLAD(s)+SP+SuperGlue+g2o & 0.956 & 0.934 & 0.693 & 0.433 & 0.967 & 0.942 & 0.701 & 0.440 \\
    Patch-NetVLAD(s)+R2D2(K)+ ICP &  \textbf{0.964} &	0.931 &	0.876 &	0.464 &	0.968 &	 \textbf{0.944} &	0.911 &	0.507 \\
    PNTR with SP+SuperGlue & 0.952 & 0.919 & 0.863 & 0.451 & 0.968 & 0.942 & 0.905 & 0.502 \\
    PNTR with R2D2(K) &	0.961 &	 \textbf{0.938} &	 \textbf{0.906} &	 \textbf{0.503} &	 \textbf{0.969} &	0.943 &	 \textbf{0.913} &	 \textbf{0.509} \\
    PNTR with LoFTR & 0.959 &	0.936 &	0.895 &	0.472 &	 \textbf{0.969} &	 \textbf{0.944} &	 \textbf{0.913} &	0.508 \\ 

    \hline
    \end{tabular}%
  \label{tab:pipeline_metrics_all}%
\end{table*}

\textbf{Localization Quality Estimation.}
The experiments were held on the HPointLoc dataset. All the methods we used were already pre-trained. 
According to Fig. \ref{fig:pipeline} and Eq.\ref{eq:retrieval}, the image retrieval method searches for the closest similar image. After that, we optimize the relative pose of the retrieved image to query image (Eq.\ref{eq:pose}). Knowing the absolute pose of the image from the database, we get the final pose of the query. The quality metrics we used are generally recognized and observable on the benchmark \cite{cvpr}. 
Accuracy metrics of true localizations with the optimization step on the HPointLoc-Val dataset are given in Table \ref{tab:image_retrieval_methods_loc}. Pose optimization in this case is not accomplished: in other words, the result pose is equivalent to the database image that was accepted as the most similar.
From Table \ref{tab:image_retrieval_methods_loc}, we can see that at all translation error thresholds the best method is Patch-NetVLAD with speed up configuration (s). The angle error should be taken into account secondary since the frame pose is not optimized.

The localization results of different methods on the HPointLoc-Val dataset are given in the Table~\ref{tab:pipeline_metrics_val}. SuperPoint is abbreviated as SP. As we can see, the highest quality for image retrieval, keypoints matching, camera pose optimization is for the proposed PNTR approach with the R2D2 feature extraction and matching based on procedure form the Kapture tool (symbol `K'). 

The localization results of different methods on the HPointLoc-All are given in Table \ref{tab:pipeline_metrics_all}. We can see that all metrics are worse in comparison with the evaluation on the HPointLoc-Val dataset. This directly follows from the degradation of the image retrieval methods.

\begin{figure}[t]
    \centering
    \includegraphics[width=1.0\textwidth]{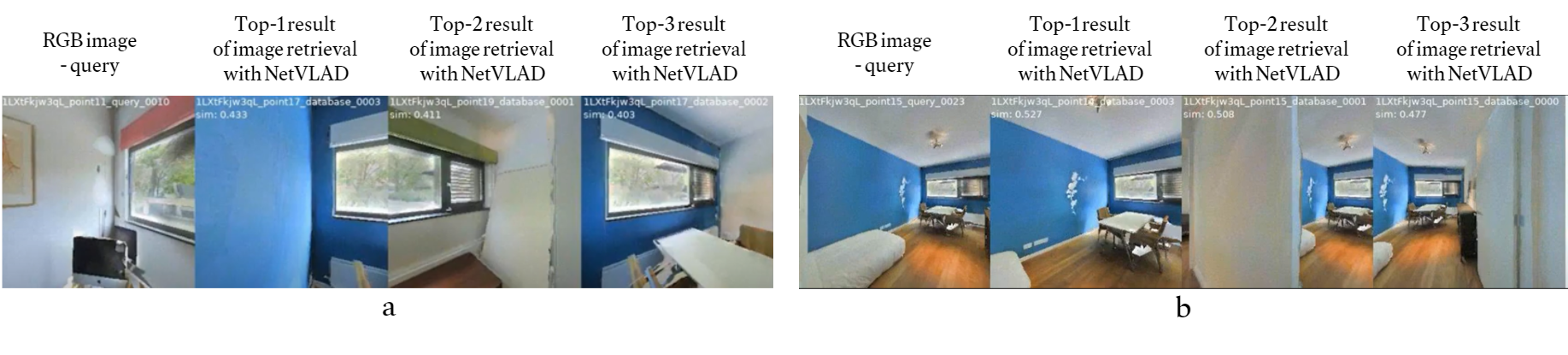}
    \caption{
    Typical problems that sometimes arise in all image retrieval methods (for example, the NetVLAD) on the HPointLoc Dataset: a — the selection of an image that looks like a query occurs with an error caused by the lack of explicit usage of semantics (information about the presence of objects, about their color, etc.); b — the selected top-1 image contains the same scene as the query image, but is more than 1 meter away from it and generates a localization error.
    }
    \label{fig:common_errors}
\end{figure}

In the course of the experiments, we detect problems that sometimes arise in all of the considered image retrieval methods. In Fig. \ref{fig:common_errors}, they are shown using the NetVLAD method as an example and demonstrate the need to explicitly use semantics and the complexity of choosing a top-1 image containing the same scene as the query image, but the corresponding camera is located at a considerable distance from the camera for images from the database.

\begin{table}[t]
    \caption{Average execution time  of different localization stages, sec}
    \centering
    \tiny
    \begin{tabular}{|l|l|l||l|l|l|p{1cm}|}
    \hline
        \multicolumn{3}{|c||}{Image        retrieval} & \multicolumn{2}{|c|}{Camera Pose Optimization} \\ \hline
        Method & Embedding extr. & Embedding match. & Method & Avg. time \\ \hline \hline
        NetVLAD & 0.00698 & - & g2o & 0.00074 \\ \hline
        AP-GeM & 0.01542 & 0.00013 & ICP & 0.00158 \\ \hline
        HF-Net & 0.09049 & - & Teaser++ & 0.00826 \\ \hline
        Patch-NetVLAD(s) & 0.01843 & 0.1039 &  &  \\ \hline
        Patch-NetVLAD(p) & 0.26415 & 2.28863 &  &  \\ \hline \hline
        \multicolumn{3}{|c||}{Local feature extraction and matching} & \multicolumn{2}{|c|}{Overall PNTR approach} \\ \hline
        Method & Feature extr. & Feature match. & Method & Avg. time \\ \hline \hline
        ORB+DBoW2 & 0.00407 & 0.00013 & PNTR with SuperPoint+SuperGlue & 0.13565 \\ \hline
        ORB+FBoW & 0.00092 & 0.00007 & PNTR with R2D2 & 0.19207 \\ \hline
        SuperPoint+SuperGlue & 0.0025 & 0.00256 & PNTR with LoFTR & 0.15189 \\ \hline
        R2D2(K) & 0.0585 & 0.00298 &  &  \\ \hline
        LoFTR & 0.0213 & - &  &  \\ \hline
    \end{tabular}
    \label{tab:inference_time}
\end{table}

\textbf{Time Performance Estimation.}
Average execution time of the main stages of localization implemented in the proposed framework is shown in Table~\ref{tab:inference_time}. The performance was evaluated on a workstation with NVidia RTX2080Ti 11Gb GPU, AMD Threadripper 1900X (8 cores, 3.8GHz) CPU, 64Gb RAM.

It should be noted that the Patch-NetVLAD (s) configuration is more than 20 times faster than the Patch-NetVLAD (p) configuration, and their quality metrics are almost the same. NetVLAD leads in speed, which is 10 times faster than the almost identical in quality HF-Net method and than the significantly better Patch-NetVLAD (s).

Among the considered methods of feature extraction, the ORB FBoW method is the leader, which is almost five times faster than the classical ORB + DBoW2 method, but is significantly inferior in quality. LoFTR, the one-stage method for extracting and matching points, has the highest performance among neural network approaches, which slightly exceeds the total performance of the SuperPoint+SuperGlue method combination.
Among the considered optimization methods, the fastest is the g2o method, which is twice as fast as ICP and almost 10 times faster than TEASER++.
The total performance of the most accurate groups of methods, except the slow Patch-NetVLAD (p), is 5-10 FPS, which indicates the potential for their practical use in a parallel global localization stream to solve the problem of loop closure in SLAM methods.

\section{Conclusions}

In this work, we have shown that the proposed HPointLoc dataset allows us to visually assess the quality and performance of various approaches for solving the place recognition problem in a photorealistic indoor environment using RGB-D images. 
This is achieved primarily by creating a regular grid of ``Points'' during dataset preparation.
This can be done for any real indoor environment whose model can be imported into the Habitat simulator.

On the small subset HPointLoc-Val, localization methods exhibit similar behavior, so it is sufficient to use it for quick evaluation and comparison of algorithms. The experiment results have revealed the limitations of the considered localization methods. It leads to the need of explicitly taking into account the semantics of the scene and the importance of a correct interpretation of the localization error when using one or the other image retrieval method.

State-of-the-art results were obtained for the proposed PNTR approach, which showed the highest quality on the developed dataset, but is rather slow and provides an image processing speed of 5 FPS. Nevertheless, it can be still used for loop detection in SLAM algorithms in a parallel stream to the main visual tracking procedure.

%
%
\bibliographystyle{splncs04}
\bibliography{ms}
\end{document}